\documentclass[sigconf,authorversion]{acmart}

\usepackage{todonotes}
\usepackage{nicefrac}

\newcommand{\approach}{\textit{StreetToPerson}}

\newcommand{\popRank}{\textbf{PopRank}}
\newcommand{\relRank}{\textbf{RelRank}}
\newcommand{\tagMe}{\textbf{TagMe}}

\copyrightyear{2022}
\acmYear{2022}
\acmConference[WWW '22 Companion] {Companion Proceedings of the Web Conference 2022}{April 25--29, 2022}{Virtual Event, Lyon, France.}
\acmBooktitle{Companion Proceedings of the Web Conference 2022 (WWW '22 Companion), April 25--29, 2022, Virtual Event, Lyon, France}
\acmPrice{15.00}
\acmISBN{978-1-4503-9130-6/22/04}
\acmDOI{10.1145/3487553.3524267}

\begin{document}

\title{Linking Streets in OpenStreetMap to Persons in Wikidata}

\author{Daria Gurtovoy}
\affiliation{%
 \institution{Universität Bonn}
 \country{Germany}}
 \email{s7dagurt@uni-bonn.de}

\author{Simon Gottschalk}
\orcid{0000-0003-2576-4640}
\affiliation{%
  \institution{L3S Research Center, Leibniz Universität Hannover}
  \country{Germany}}
\email{gottschalk@L3S.de}

\begin{abstract}
Geographic web sources such as OpenStreetMap (OSM) and knowledge graphs such as Wikidata are often unconnected. An example connection that can be established between these sources are links between streets in OSM to the persons in Wikidata they were named after. This paper presents \approach{}, an approach for connecting streets in OSM to persons in a knowledge graph based on relations in the knowledge graph and spatial dependencies. Our evaluation shows that we outperform existing approaches by $26$ percentage points. In addition, we apply \approach{} on all OSM streets in Germany, for which we identify more than $180,000$ links between streets and persons.
\end{abstract}



\maketitle

\section{Introduction}
\label{sec:introduction}

Streets are often named after famous or distinguished individuals who may or may not have a direct connection to the specific location. While geographic data sources such as OpenStreetMap~(OSM)\footnote{OpenStreetMap, OSM and the OpenStreetMap magnifying glass logo are trademarks of the OpenStreetMap Foundation, and are used with their permission. We are not endorsed by or affiliated with the OpenStreetMap Foundation.} contain information about streets across the globe and knowledge graphs such as Wikidata contain information about famous individuals and their relations, these two worlds are often not connected. This makes it challenging to connect streets to whom they are named after.

For example, consider the street named ``Wilhelmstraße'' in the German capital Berlin. After the removal of the suffix ``straße'' (German for ``street''), only the term ``Wilhelm'' remains. This term can potentially be linked to a variety of persons, be it via their first name (e.g., Wilhelm Busch), their second name (e.g., Friedrich Wilhelm I.) or their last name (e.g., Paul Wilhelm). With the help of Wikidata, we can infer that Friedrich Wilhelm I. was born in Berlin and was a historically important monarch. Both factors may lead to the correct selection of Friedrich Wilhelm I. for the Wilhelmstraße.

The provision of links between streets in OSM to persons in Wikidata can support several use cases including: (i)~Bridging the gap between OSM and knowledge graphs: By bringing OSM and Wikidata together, they will mutually benefit from their strengths~\cite{tempelmeier2020}. Knowledge graphs such as WorldKG~\cite{worldkg} and the Nuremberg Address Knowledge Graph~\cite{BrunsTCPXS21} are examples of such efforts. (ii)~Providing data for touristic applications: Tourists are often interested in exploring the history behind a place that can often be told by people who have a special significance in that place. An existing example is an application for studying the backgrounds behind the \textit{Stolpersteine} in Germany, which are dedicated to victims of the Nazi regime\footnote{\url{https://stolpersteine.wdr.de/web/en/}}.
(iii)~Enabling cultural analyses: With street-to-person links available, analyses are enabled about the motives behind naming streets, potentially revealing discrimination of women~\cite{ouali2021women} and the detection of quarters dedicated to specific groups of persons~\cite{almeida2016}.

In contrast to named entity linking approaches on text, no context information is available additional to the street's name and location. Therefore, the task of linking streets in OSM to persons in Wikidata lies in identifying the correct person based on features such as the person's popularity and the person's relatedness to the street's location. We propose \approach{}, which first builds an index for retrieving potential candidate persons given a street name. For each such candidate person, we extract a set of features denoting specific characteristics of that person, such as its popularity and its spatial relatedness to the given street, and classify it as correct or not.

In our evaluation, we first show \approach{}'s superiority over existing baselines using a ground truth extracted from German streets in Wikidata, and we apply \approach{} to German streets in OSM. \approach{} reaches a precision of $0.95$ on the ground truth and identifies more than $180,000$ street-to-person relations.

The remainder of this paper is structured as follows: In Section~\ref{sec:related_work}, we present related work. Then, in Section~\ref{sec:approach}, we describe our approach and evaluate it in Section~\ref{sec:evaluation}. Finally, we provide a conclusion in Section~\ref{sec:conclusion}.

\begin{figure*}[ht]
\centering
\includegraphics[width=0.8\textwidth]{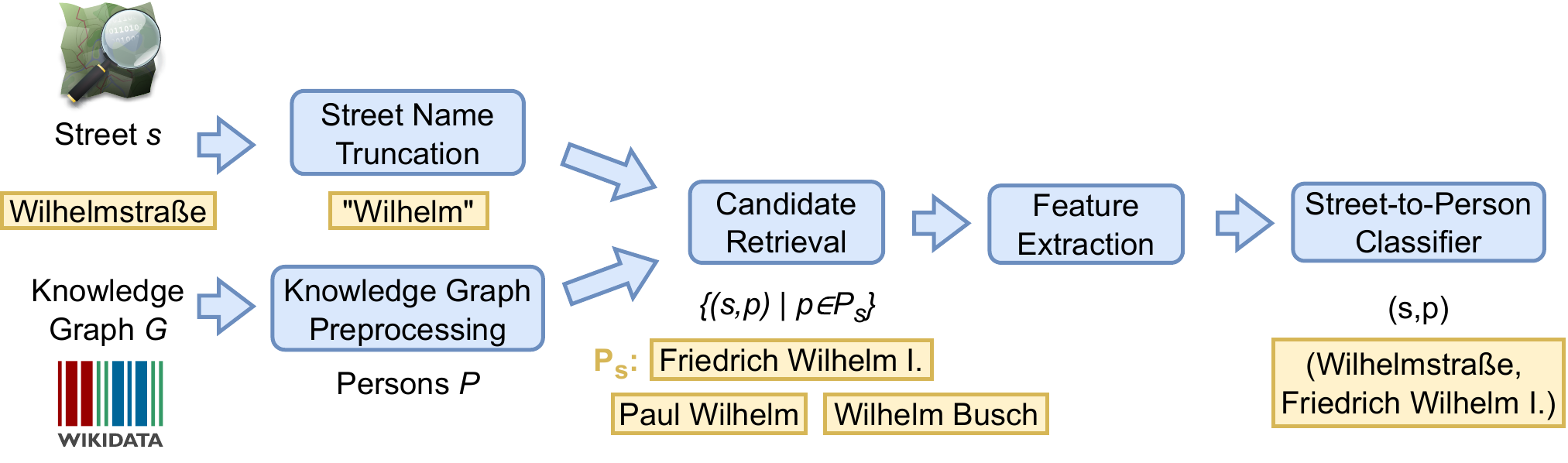}
\caption{Overview of \approach{}. Yellow boxes show example values.}
\label{fig:pipeline}
\end{figure*}
\section{Related Work}
\label{sec:related_work}

Linking streets to persons is related to named entity linking and connecting geographic data sources with knowledge graphs.

\subsection{Named Person Entity Linking}

Most related to this paper is the work by Almeida et al.~\cite{almeida2016}, a ranking-based approach to connect streets to the persons they were named after based on a manually defined relevance score. Users then confirm the generated street-to-person pairs in a web interface. Geiss et al. \cite{geiss2016} link mentions of persons in a text document to Wikidata using a network of candidate persons. In general, street-to-person linking can be considered a variation of named entity linking to Wikidata \cite{moeller2022}.

\subsection{Connecting OSM with Knowledge Graphs}

The task of connecting geographic data sources such as OSM to knowledge graphs has been addressed from different perspectives. Typically, approaches aim at establishing identity links between the different representations of geographic entities and concepts in these sources. For example, \cite{tempelmeier2020} proposes a pipeline for link discovery between OSM, Wikidata and DBpedia based on OSM tags, \cite{dsouza2021towards} aligns the schema between these sources using an adversarial
classifier, and osm2rdf converts the whole OpenStreetMap data to RDF triples \cite{bast2021efficient}. In contrast to these approaches, our task deals with persons, a class of entities not present in OSM.

\section{Approach}
\label{sec:approach}

The goal of this paper is to link streets to the correct person whom they are named after. Formally, we develop a \textit{street-to-person mapping function} as follows:

\begin{definition}
\textbf{Street-to-person mapping function.} Given a street $s$ and a set of persons $P$, create a street-to-person mapping function $f(s) \mapsto P$ that identifies the person $p \in P$ after whom the street is named.
\end{definition}

Figure~\ref{fig:pipeline} illustrates the approach and its components. First, the name of the given street $s$ is truncated to potentially only contain a person's name. This term is then used in a candidate generation step to retrieve a set of candidate persons $P_s$ out of a set of persons $P$ extracted from Wikidata. For each street-candidate-pair $(s,p)$, we extract various features and use a street-to-person classifier that determines the candidate with the highest probability as the correct person for the given street, where the street can be of any type of street in any region. In the example given in Figure~\ref{fig:pipeline}, the street "Wilhelmstraße" is truncated to "Wilhelm", for which the preprocessed knowledge graph returns the candidates ``Paul Wilhelm'', ``Wilhelm Busch'' and ``Friedrich Wilhelm I.'' and others. After extracting all candidate features, the classifier determines the latter to be the best candidate for the given street.

The implementation of this pipeline as well as the generated links are available on GitHub\footnote{\url{https://github.com/d-gurtovoy/streetnameLinks}}.

\subsection{Knowledge Graph Preprocessing}

\label{02-kg-preprocessing}

The data required to retrieve candidates and their features is taken from Wikidata. For efficient access to relevant information, we create the following data sets from Wikidata in a preprocessing step\footnote{To extract these data sets, we process the Wikidata dump using the \textit{qwikidata} library (\url{https://qwikidata.readthedocs.io/}).}:

\begin{itemize}
    \item The \textit{ground truth} for training and evaluation contains known streets in Germany that are named after a person. We provide details about this dataset in Section~\ref{subsec:data}.
    \item The \textit{person index} takes a term as input and returns the Wikidata IDs of all people that match it. This person index contains over 4 million names of over 9 million persons.
    \item The \textit{person occupation index} contains all occupations of a person (e.g., monarch or writer).
    \item The \textit{person location index} contains relevant locations such as a person's birthplace.
    \item The \textit{spatial dependency database} denotes containment relations between locations (e.g., Berlin is located in Germany).
\end{itemize}

\begin{figure}[ht]
\centering
\includegraphics[width=\columnwidth]{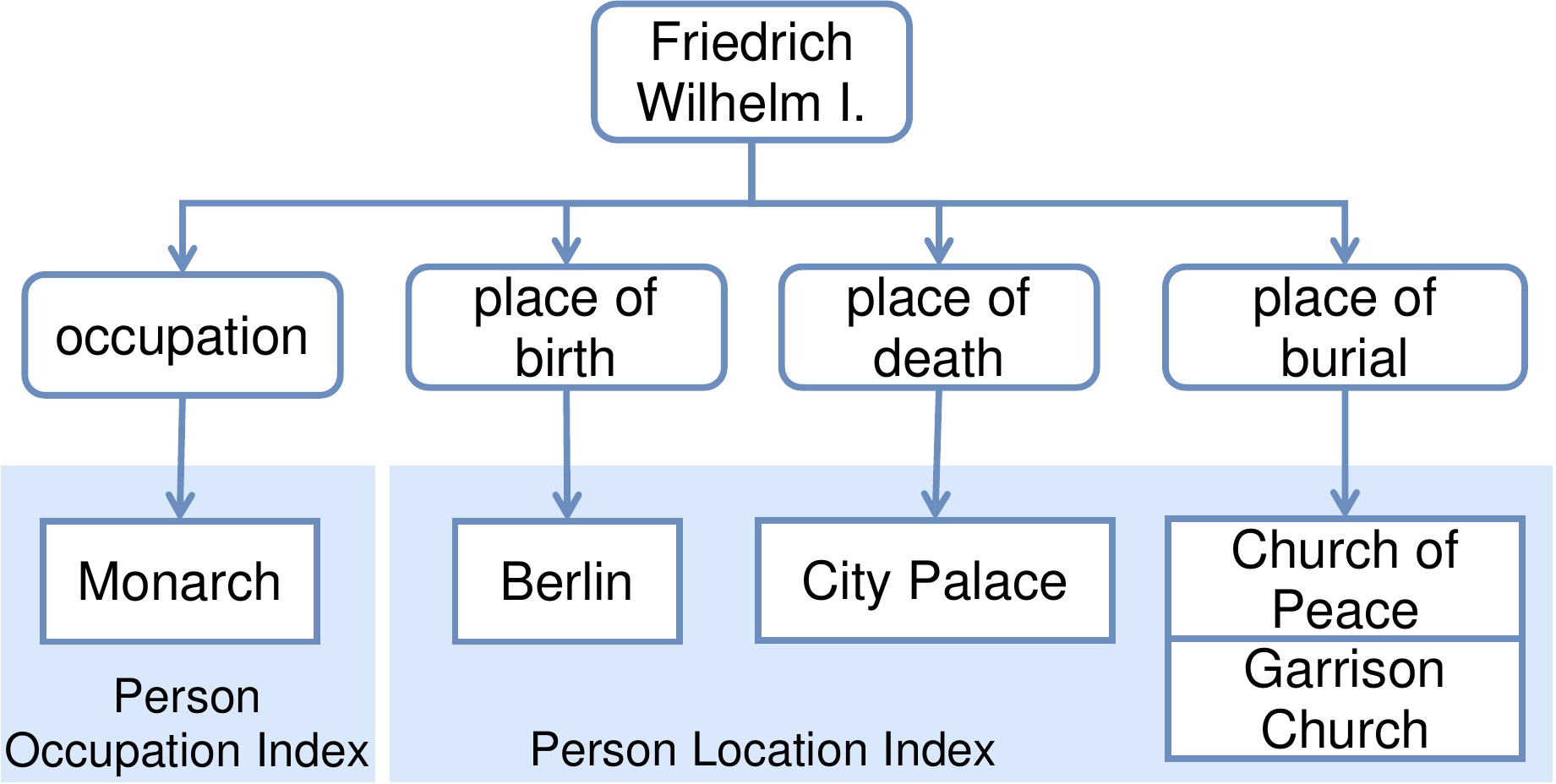}
\caption{Relations of Friedrich Wilhelm I. from Wikidata.}
\label{fig:person_locations}
\end{figure}

Figure~\ref{fig:person_locations} shows relations of Friedrich Wilhelm I. retrieved from the person occupation index and the person location index.\footnote{Over time,  Friedrich Wilhelm I. has been buried in different churches.}

\subsection{Street Name Truncation}
When analysing a street name, it is essential to differentiate between the part that maps to a person and street affixes (i.e., prefixes and suffixes), such as ``street'', ``road'', and ``avenue''. Given an extensive list of common street affixes, they can be removed from the street name so that only the person name remains, which can then be used to look up matches in the person index database. To achieve this, we gathered the streets of the ground truth and removed all names and aliases from the persons associated with them. 
The remaining parts of the street names are then manually split into prefixes and suffixes and corrected when necessary. The result is a set of 80 German suffixes and 34 prefixes which we make available\footnote{\url{https://github.com/d-gurtovoy/streetnameLinks/tree/master/data/affixes}}.
Using this set of affixes, the name of a street $s$ is truncated so that street affixes are removed and the part that potentially maps to a person name remains.

\subsection{Candidate Retrieval}
The truncated street name is used to query the person index, which returns the Wikidata IDs of matching candidates ${p \in P_s}$.
For instance, the term ``Wilhelm'' would return the ID of ``Friedrich Wilhelm I.'' but also of ``Wilhelm Busch'' among others.

\subsection{Feature Extraction}
After truncating the street name and retrieving candidates, we extract characteristics of the person $p$ and the spatial relations between $p$ and the street $s$ that serve as features for training a binary classifier. 
The following $30$ features are extracted for every street-candidate-pair $(s,p)$:

\begin{itemize}
    \item \textit{Link count}: The number of links pointing to $p$ in the German Wikipedia. If the candidate does not exist in Wikipedia but only in Wikidata, this feature value is set to $0$.
    \item \textit{Name}: These four binary features show which part of the person's name is contained in the name of $s$. It can be its full, first or last name or an alias.
    \item \textit{Occupations}: From the person occupation index, we gather $20$ of the most common occupations of people that German streets were named after and introduce $20$ binary features to mark if the person $p$ held any of these positions.
    \item \textit{Spatial relations}: We introduce five numerical features representing whether there is a spatial relation between $p$ and $s$: ``born'', ``died'', ``buried'', ``educated at'', and ``work location''.
\end{itemize}

The \textit{spatial relations} feature values represent to which extent the street and the person's related location are contained within each other. These feature values range from $0$ to $1$, with $0$ meaning the given location is either unknown or lies outside of Germany and $1$ if the location is the street itself. If a person is related to multiple locations for the same relation, we take the location with the highest score.

\begin{figure}[t]
\centering
\includegraphics[width=0.85\columnwidth]{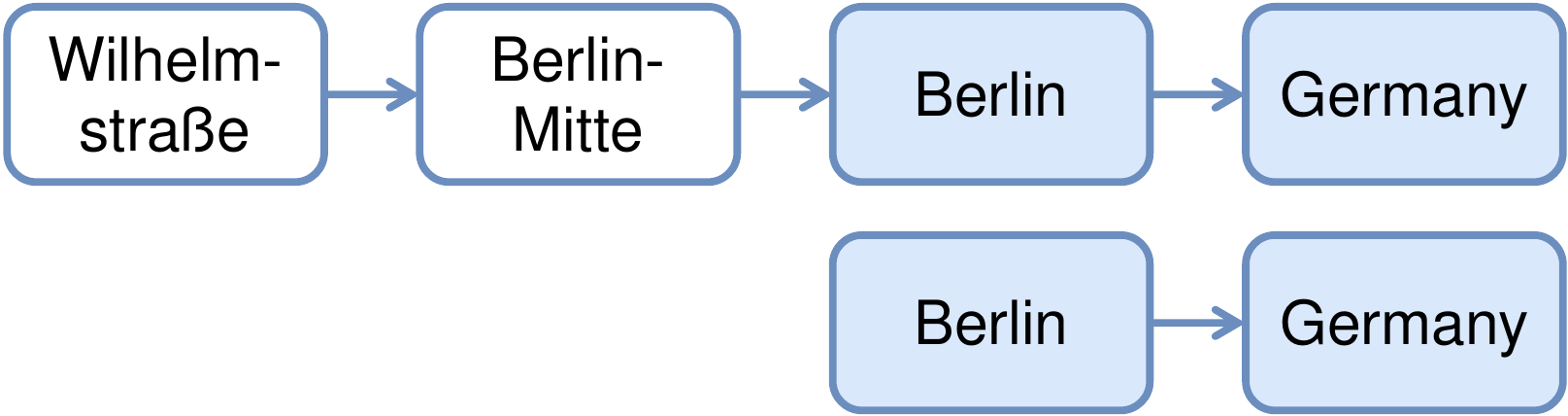}
\caption{Example of the spatial dependencies between the street ``Wilhelmstraße'' and the city Berlin. Arrows denote ``located in'' relations (e.g., Berlin is located in Germany).}
\label{fig:region}
\end{figure}

Figure~\ref{fig:region} illustrates an example of computing the containment of the street ``Wilhelmstraße'' and Berlin, the birthplace of Friedrich Wilhelm I. We create a containment chain for both locations from the spatial dependency database and check their overlap. In this example, the ``born'' relation feature is set to $\nicefrac{2}{4}=0.5$.

\section{Evaluation}
\label{sec:evaluation}

This section introduces the data used for training and testing and evaluates \approach{} against three baselines.

\subsection{Data}
\label{subsec:data}

We require a dataset of known person-to-street links to train our approach and evaluate. For a few cases, Wikidata and OSM provide this information. Wikidata contains $4,799$ German streets connected to a person via Wikidata's ``named after'' property\footnote{\url{https://www.wikidata.org/wiki/Property:P138}}. We use these street-to-person relations as positive examples for training our model. As negative examples, we retrieve up to $50$ candidate persons of these streets with the highest link count each.

OSM provides a tag called ``name:etymology''\footnote{\url{https://wiki.openstreetmap.org/wiki/Key:name:etymology}}, which links streets to persons in Wikipedia\footnote{We use Wikidata \textit{sitelinks} to link persons in Wikipedia to Wikidata}. This key provides $19,592$ street-to-person relations involving German streets.

\subsection{Baselines}

We compare \approach{} to three different baselines:

\begin{itemize}
    \item TagMe~\cite{tagme} (\tagMe{}): \tagMe{} is a traditional entity linking approach on short text fragments. As it may link a street name to the actual street entity and not the person it is named after, we feed \tagMe{} with the street name after applying street name truncation.
    \item Popularity Ranking (\popRank{}): In a simpler version of \approach{}, we replace the street-to-person classifier by taking the person with the highest link count.
    \item Relevance Ranking~\cite{almeida2016} (\relRank{}): The approach by Almeida et al. described in Section~\ref{sec:related_work} is, to the best of our knowledge, the only existing approach for street-to-person linking. As \relRank{} does not limit its results to person entities, we consider two configurations: \relRank{} (all entities) and \relRank{} (person entities), where we only consider links to person entities.
\end{itemize}

\subsection{Evaluation of the Classification}

We evaluate \approach{}'s classification performance on the Wikidata ground truth described in Section~\ref{subsec:data} using $10$-fold cross-validation. Results are shown in Table~\ref{tab:evaluation}. \approach{} clearly outperforms the other approaches and reaches a precision of $0.95$, $26$ percentage points more than the second-best baseline, \popRank{}. The low recall of \relRank{} can be explained by its limited number of German affixes (``straße'' and ``weg''), the low amount of features and the manually created relevance criterion.

\begin{table}[ht]
\centering
\begin{tabular}{@{}lrrr@{}}
\toprule
 & \textbf{Precision} & \textbf{Recall} & \textbf{F1 Score} \\ \midrule
\begin{tabular}[c]{@{}l@{}}\tagMe{}\end{tabular} & 0.49 & 0.45 & 0.47 \\
\popRank{} & 0.69 & 0.66 & 0.67 \\
\begin{tabular}[c]{@{}l@{}}\relRank{} (all entities)\end{tabular} & 0.08 & 0.08 & 0.08 \\
\begin{tabular}[c]{@{}l@{}}\relRank{} (person entities)\end{tabular} & 0.35 & 0.11 & 0.17 \\
\textbf{\approach{}} & \textbf{0.95} & \textbf{0.91} & \textbf{0.93} \\ \bottomrule
\end{tabular}
\caption{Evaluation of the classification for \approach{} and the selected baselines.}
\label{tab:evaluation}
\end{table}

\subsection{Application on OpenStreetMap}

In a second step of the evaluation, we demonstrate how \approach{} can be used for identifying street-to-person relations between OpenStreetMap and Wikidata that are not yet contained in these sources. To this end, we apply \approach{} to the whole of Germany and two selected German states -- the most populated state, North Rhine-Westphalia (NRW), and the least populated state, Bremen. From OSM, we select all streets contained in these regions and then apply \approach{} as depicted in Figure~\ref{fig:pipeline}. Table~\ref{tab:evaluation_osm_stats} shows the results: For more than half of the $219,768$ streets in NRW (Bremen: $2,504$ of $6,733$ streets), the person index returned at least one candidate person. After applying the street-to-person classifier, $28,857$ street-to-person relations were identified (Bremen: $896$). In the whole of Germany, more than $180,000$ street-to-person relations are identified.
This table also emphasizes the difficulty of identifying the correct person candidate: For $669,304$ streets in Germany, more than $16$ million candidate persons are found. An example of a false classification is the street "Grevesmühlweg" in Bremen which is wrongly assigned to Maria Grevesmühl but should be assigned to her father Hermann Grevesmühl -- both musicians who were born and died in Bremen, thus highly similar.

\begin{table}[ht]
\centering
\begin{tabular}{@{}lrrr@{}}
\toprule
\textbf{Number of} & \multicolumn{1}{r}{\textbf{Bremen}} & \textbf{NRW} & \textbf{Germany} \\ \midrule
Streets & $6,733$ & $219,768$ & $1,321,464$ \\
\ \ \ with candidate persons  & $2,504$ & $110,968$ & $669,304$ \\
Candidate persons & $47,659$ & $2,675,761$ & $16,165,454$ \\
Street-to-person relations & $896$ & $28,857$ & $183,022$ \\ \bottomrule
\end{tabular}
\caption{Number of street-to-person relations identified for OSM streets in Germany and two of its states.}
\label{tab:evaluation_osm_stats}
\end{table}

Finally, we measure the correctness of the identified street-to-person pairs based on a subset of the streets in Table~\ref{tab:evaluation_osm_stats}, which are assigned to a person in OSM through the \texttt{name:etymology} key. Table~\ref{tab:evaluation_osm} shows the results on this subset of $45$ (Bremen) and $1,352$ streets (NRW). The results show a precision of $0.90$ for NRW and $0.94$ for Bremen which confirms the generalizability of our approach from Wikidata to OSM.

\begin{table}[ht]
\centering
\begin{tabular}{@{}lrr@{}}
\toprule
 & \multicolumn{1}{l}{\textbf{Bremen}} & \multicolumn{1}{l}{\textbf{NRW}} \\ \midrule
\textbf{Precision} & $0.94$ & $0.90$ \\
\textbf{Recall} & $0.64$ & $0.61$ \\
\textbf{F1 Score} & $0.76$ & $0.73$ \\ \bottomrule
\end{tabular}
\caption{Evaluation of \approach{} on the OSM ground truth in two German states.}
\label{tab:evaluation_osm}
\end{table}

\section{Conclusion and Future Work}
\label{sec:conclusion}

In this paper, we have presented \approach{}, an approach for connecting streets in OpenStreetMap to those persons in Wikidata whom they were named after. Through a combination of knowledge graph features and spatial features, \approach{} precisely identifies new relations. In future work, we plan to enrich the feature space through the utilisation of graph-based embeddings and to extend our approach to further countries and other relations between geographic entities in OpenStreetMap and knowledge graph entities.


\begin{acks}
This work was partially funded by the Federal Ministry of Education and Research (BMBF), Germany under ``Simple-ML'' (01IS18054) and the DFG, German Research Foundation, under ``WorldKG'' (424985896).
\end{acks}

\bibliographystyle{ACM-Reference-Format}
\bibliography{main}

\end{document}